\pdfoutput=1
\documentclass[journal]{IEEEtran}

\usepackage[utf8]{inputenc} 
\usepackage[T1]{fontenc}    
\usepackage{hyperref}       
\usepackage{url}            
\usepackage{booktabs}       
\usepackage{amsfonts}       
\usepackage{nicefrac}       
\usepackage{microtype}      
\usepackage{lipsum}
\usepackage{graphicx}        
\usepackage{multicol}        
\usepackage{multirow}

\title{Relevant features for Gender Classification in NIR Periocular Images}

\author{
  Ignacio Viedma, \\ Engineering Science Department, Universidad Andres Bello, Chile.\\ \texttt{i.viedmaescalona@uandresbello.edu} \\
  Juan Tapia,\\ Universidad Tecnologica de Chile - INACAP, Chile.\\ \texttt{j\_tapiaf@inacap.cl} \\
  Andres Iturriaga \\ Universidad de Chile, Program of Biostatistics, School of Public Health, Faculty of Medicine, Chile. \\ \texttt{aiturriagaj@gmail.com} \\
  Christoph Busch \\ da/sec - Biometrics and Internet Security Research Group, Hochschule Darmstadt, Germany. \\ \texttt{christoph.busch@h-da.de} \\
   \vspace{0.4cm}
   \scriptsize{\textbf{***This paper is a preprint of a paper accepted by IET Biometrics and is subject to Institution of Engineering and Technology Copyright. When the final version is published, the copy of record will be available at the IET Digital Library***.}}
}

\begin{document}
\maketitle

\begin{abstract}
Most gender classifications methods from NIR images have used iris information. Recent work has explored the use of the whole periocular iris region which has surprisingly achieve better results. This suggests the most relevant information for gender classification is not located in the iris as expected. In this work, we analyze and demonstrate the location of the most relevant features that describe gender in periocular NIR images and evaluate its influence its classification. Experiments show that the periocular region contains more gender information than the iris region. We extracted several features (intensity, texture, and shape) and classified them according to its relevance using the XgBoost algorithm. 
Support Vector Machine and nine ensemble classifiers were used for testing gender accuracy when using the most relevant features. The best classification results were obtained when 4,000 features located on the periocular region were used (89.22\%). 
Additional experiments with the full periocular iris images versus the iris-Occluded images were performed. 
The gender classification rates obtained were 84.35\% and 85.75\% respectively. We also contribute to the state of the art with a new database (UNAB-Gender). From results, we suggest focussing only on the surrounding area of the iris. This allows us to realize a faster classification of gender from NIR periocular images.
\end{abstract}
\maketitle


\section{Introduction}
\label{sec1}

Periocular biometrics have recently attracted much attention since it allows improvement of the robustness of face or iris biometric recognition. The periocular region is the area surrounding the eye, and it is generally considered one of the most discriminative regions of the face \cite{Merkow2010}. It has been shown that the periocular region itself can be used for recognition \cite{Alonso-FernandezBigun2016}. It can also help with iris recognition when the inherent biometric content within the source images of poor quality. This suggests that periocular features could have a potential for soft biometric classification.

Recently, Soft-biometrics studies have proposed the use of iris information for estimating demographic information such as gender, ethnicity, age, and emotions. \cite{AbdelAlim2002, Merkow2010, Popescu-Bodorin2011, Qiu2006, Ricanek2011,  Alexandre2010, Tapia2013, RattaniReddyDerakhshani2017, SinghNagpalVatsaEtAl2017, Eidinger2014Age, Rattani2017CNN}. 
Gender information, in particular, can improve biometric recognition administration systems, since it allows reduced search time \cite{JuanE.Tapia2014}. This is especially important in countries such as India, China and others which have large populations \cite{CANPASS1996, Daugman2009, UIP2014}. For instance, if gender is determined then the average search time may be halved. 


Gender information may also be useful when people are not recognized but is attempting to gain entry to a restricted zone. Another possible use of the gender information arises in social settings, where it may be useful to screen entry to an area based on the gender of a person, while not recording identity \cite{TapiaPerezBowyer2016}. Gender classification is also crucial for banking transactions from cellular phones applications, demographic information collection, marketing research, and real-time digital marketing \cite{Han2014, Perez2012, Tapia2013}. Therefore, this information is essential to manage iris recognition systems better.

Most gender classification methods reported in the literature use the entire face as Region of Interest (ROI) \cite{Merkow2010}. However, the performance of these methods suffers greatly when portions of the face are occluded. Recent studies have narrowed the principal region for biometric identification to the surrounding eye area. This area is known as the periocular region \cite{ParkRossJain2009}.

Most periocular algorithms work in a holistic fashion. They usually define the rectangle that contains the whole eye area as an ROI. The full rectangular area is then used for feature extraction. Such a holistic approach implies that some components that are not relevant for identity recognition, like hair or glasses, may erroneously bias the recognition process. Furthermore, features may not be equally discriminative in all parts of the periocular regions.

Periocular biometrics techniques have mainly focused on texture features. These features are calculated from a single periocular image. 
Therefore, can not efficiently handle variations introduced by movements of the eyeball and eyelid. The performance further suffers if we add variations in pose and illumination.

Hollingsworth et al. \cite{HollingsworthDarnellMillerEtAl2012} identified some of the ocular elements that are found most useful for periocular recognition. The relevance of those elements depends on the technology used for capturing the image.
For Near-Infra-Red (NIR) images the most influential elements are the tear duct and lower lash. The skin and cheek are the least relevant. When using Visual Spectrum (VIS) images, on the contrary, skin texture is shown to be more important as well as the blood vessels and the brown region. 
Those results agree with assumptions made in the literature indicating NIR illumination is more suitable for iris texture recognition \cite{Daugman2004} while VIS illumination performs better for analyzing the periocular region.
Several studies have been carried out using automatic algorithms with similar results.


We have previously demonstrated that gender classification based on iris information (NIR images) achieves similar results as classical method using facial biometric features \cite{TapiaPerezBowyer2016,Tapia2017}. 

In this paper, we analyze the relevance of the different areas around the eye. Using the same periocular NIR images used in iris-identification systems, we investigated the relevance of the iris on classifying gender in an effort to identify the best feature extraction techniques to increase gender classification rates. We also question whether the fusion of iris and periocular information is complementary. Further we contribute two new gender labeled databases, which we captured for this study. 

This paper is organized as follows. In Section 2, we review the state of the art in gender classification. In Section 3, we describe the proposed method and present the feature extraction techniques used. In section 4, we provide details about the datasets used in our experiments. In Section 5, we describe the experiments and results, and in Sections 6 and 7, we present the discussion and conclusions, respectively.

\section{State of the Art}

\begin{table*}[h]
\scriptsize
\centering
\caption{Summary of gender classification using eyes: I = Iris Images, P = Periocular Images, L = Left and R = Right, FEM = Feature Extraction Method, CCR= Correct Classification Rate.} \label{summary}
\begin{tabular}{|c|c|c|c|c|c|c|c|}
\hline
Paper & I/P/F & Source & $N^o$ of Images & $N^o$ of Subjects & Type & FEM & CCR (\%)  \tabularnewline
\hline
Thomas et al. \cite{Thomas2007} & I & Iris & 16,469 & N/A & NIR & Gabor Filter & 75.00  \tabularnewline
\hline
Lagree et al.\cite{Lagree2011} & I & Iris & 600 & 300 & NIR & vGabor Filter & 62.17 \tabularnewline
\hline
Bansal et al. \cite{Bansal2012} & I & Iris & 400 & 200 & NIR & Discrete wavelet transform & 83.60 \tabularnewline
\hline
Tapia et al. \cite{JuanE.Tapia2014} & I & Iris & 1,500 & 1,500 & NIR & LBP & 91.00 \tabularnewline
\hline
Costa et al. \cite{Costa-Abreu2015} & I & Iris & 1,600 & 200 & NIR & Shape+Texture Features, Gabor Filter & 89.74 \tabularnewline
\hline
\multirow{2}{*}{Bobeldyk et al.\cite{BobeldykRoss2016}} & \multirow{2}{*}{I / P} & \multirow{2}{*}{Iris} & \multirow{2}{*}{3,314} & \multirow{2}{*}{1,083} & \multirow{2}{*}{NIR} & \multirow{2}{*}{BSIF} & 85.70 (P) \tabularnewline
 & & & & & & & 65.70 (I)  \tabularnewline
\hline
Merkow et al. \cite{Merkow2010} & P & Faces & 936	& 936 & VIS & LBP & 80.00 \tabularnewline
\hline
Chen et al. \cite{ChenRoss2011}	& P	& Faces	& 2,006	& 1,003	& NIR/Thermal & LBPH & 93.59 \tabularnewline
\hline
\multirow{2}{*}{\cite{Castrillon-SantanaLorenzo-NavarroRamon-Balmaseda2016}} & \multirow{2}{*}{P} & \multirow{2}{*}{Faces} & \multirow{2}{*}{3,000} & \multirow{2}{*}{1,500} & \multirow{2}{*}{VIS} & HOG,LBP,LTP & \multirow{2}{*}{92.46} \tabularnewline
 & & & & & & {{WLD,LOSIB}} & \tabularnewline
\hline
Kuehlkamp et al. \cite{KuehlkampBeckerBowyer2017} & I & Iris & 3,000 & 1,500 & NIR & LBP & 66.00 \tabularnewline
\hline
Rattani et al. \cite{RattaniReddyDerakhshani2017} & P & Faces & 572 & 200 & VIS & LBP,LPQ,BSIF & 91.60 \tabularnewline
\hline
\multirow{2}{*}{Tapia et al. \cite{Tapia2017}} & \multirow{2}{*}{I} & \multirow{2}{*}{Iris} & 10,000 unlabel & -- & \multirow{2}{*}{NIR} & {{RBM}} & 77.79  \tabularnewline
 & & & 3,000 labeled & 1,500 & & CNN & 83.00 \tabularnewline
\hline
Kuehlkamp et al. \cite{Kuehlkamp-2019} & P & Iris & 6,240 & 2005 & NIR & Hand-Crafted,CNN & 80.80 \tabularnewline
\hline
\multirow{5}{*}{This paper} & \multirow{5}{*}{P} & \multirow{5}{*}{Iris/Faces} & 952 (P-UND L) & 952 (P-UND L) & \multirow{5}{*}{NIR} & \multirow{5}{*}{{{RAW,ULBP,HOG}}} & \multirow{5}{*}{85.70}\tabularnewline
 & & & 946 (P-UND R) & 946 (Pe-UND R) & &  & \tabularnewline
 & & & 3,840 (C.E.) & 120 (C.E.) & &  & \tabularnewline
 & & & 389(UTIRIS)  & 79 (UTIRIS) & &  & \tabularnewline
 & & & 1,360(UNAB-Gender) & 136 (UNAB-Gender) & &  & \tabularnewline
This paper & & & & & &  & \textbf{89.22 (4,900)} \tabularnewline
With XgBoost & & & & & &  & \tabularnewline
\hline
\end{tabular}
\end{table*}

Gender classification initially used a variety of features extracted from facial areas. \cite{Perez2012, Tapia2013} reported the use of a feature-selection method based on mutual information and feature fusion to improve the gender classification of facial images. 

Castrillon-Santana et al. \cite{Castrillon-SantanaLorenzo-NavarroRamon-Balmaseda2016} proposed a gender-classification system that works with periocular images. This area is extracted after normalizing the face in terms of scale and rotation. Given the rough eye-location, the normalized facial image is obtained automatically after rotating, scaling, and cropping the original images. The best result was obtained using five different feature-extraction methods. Also, some studies have made use of annotated data, for the detection and segmentation of the periocular region of the eye \cite{Alonso-FernandezBigun2016}. This technique has become a research target itself.

Alonso-Fernandez et al. \cite{Alonso-FernandezBigun2016} report a survey with the most commonly used techniques based on periocular images and databases. They provide a comprehensive framework covering different perspectives, from existing databases to algorithms for the detection of the periocular region and features for recognition. Databases that were utilized include face and iris images (since the periocular area appears in such data), as well as newer databases that specifically capture the periocular area \cite{Alonso-FernandezBigun2016}.

Kumari et al. \cite{KumariBakshiMajhi2012} propose a novel approach of extracting global features from the periocular region of poor-quality grayscale images for gender classification. In their approach, global gender features are extracted using independent component analysis and evaluated using conventional neural-network techniques. The authors compare the performance, and all the best results are coming from periocular region cropped from the FERET face database  \cite{PhillipsWechslerHuangEtAl1998}. 

Studies based on NIR normalized or encoded images of the iris have explored the possibility of automatically deducing the gender of an individual \cite{Bansal2012,Lagree2011,Thomas2007, JuanE.Tapia2014, BobeldykRoss2016, TapiaPerezBowyer2016}. Previous research on this topic has extracted and used the iris region, while most operational iris biometric systems typically acquire the periocular region for processing. Very few databases have been designed specifically for gender-classification from NIR periocular research. 

Bobeldyk et al. \cite{BobeldykRoss2016} explored the correct gender-classification rate associated with four different regions from NIR iris images: the periocular region, the iris-excluded ocular region, the iris-only region, and the normalized iris-only region. They used a binarized statistical image feature (BSIF) texture operator to extract features from the regions that were previously defined. The ocular region reached a correct classification rate of 85.7\%, while the normalized or unwrapped images exhibited the worst correct classification rate at only 65\%. This suggests that the normalization process may filter out useful information.  

Kuehlkamp et al. \cite{Kuehlkamp-2019}, Very recently, analyses the influence of the feature extracted from normalized and periocular images to classify gender in a larger database. The authors used different probabilistic occlusion masking to gain insight into the problem. Results also suggest the discriminative power of the iris texture for gender is weaker than previously thought, and that the gender-related information is primarily in the periocular region. Regardless of the classifier (SVM or CNN), there is a clear and significant difference between predicting gender from periocular images and normalized iris images.
A summary of the most relevant work in this area is presented in Table \ref{summary}. 

\section{Methods}

\begin{figure}[h]
\begin{centering}
\begin{tabular}{c}
{\includegraphics[scale=0.35]{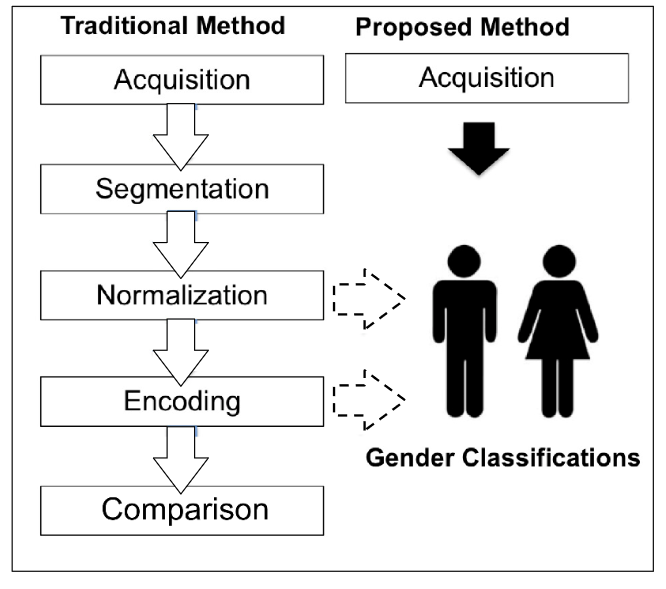}}
\end{tabular}
\par\end{centering}
\caption{ Representation of the iris-identification stages used to extract information for gender-classification. The white arrows show the traditional stages of iris recognition.  The dot-line arrows show the stages used for the traditional gender classification methods. The black arrow on the right shows the stage used in the proposed method.}
\label{fig:block}
\end{figure}

Traditional iris-identification systems have five stages: acquisition, segmentation, normalization, encoding, comparison, and decision making. Periocular NIR images can be obtained from the first stage of acquisition. See Figure \ref{fig:block}.

n this work, we directly used the acquired image from a conventional NIR sensor for extracting the features used in gender-classification. Despite other methods that require pre-processing stages such as Segmentation, Normalization, Encoding and Comparison (See Figure \ref{fig:block}).

Intensity, shape, and texture information from periocular images were used to evaluate the gender classification performance. There is no need for additional steps to achieve the soft-biometrics information. 

We used four different types of iris features obtained from periocular images to classify gender: intensity (raw image), texture features (Uniform Local Binary Patterns (ULBP)), shape (histogram of oriented gradient (HOG) and inverse HOG (IHOG). 

For all datasets the image size were $120\times160$ pixels (See Figure \ref{fig:muestras}). Along of this research, we have done a preliminary test using the original sizes of the images, but the results were very similar when the size is reduced. Since the resolution did not affect our correct classification rate we keep doing all the experiments with $120\times160$ pixels.

\subsection{Intensity images}

As intensity, we used grayscale pixel values of NIR images. Eight bits per pixel were used. The UND periocular database was captured by a regular LG 4000 NIR sensor. The UTIRIS Database \cite{HosseiniAraabiSoltanian-Zadeh2010} used a ISG Lightwise LW camera. The NIR Cross-Eyed Databases \cite{ASLC2016}, on the other hand, used a custom developed sensor that can capture simultaneously the NIR and RGB image. Therefore, the conditions and the distances while capturing images are different.

\subsection{Texture - Uniform Local Binary Patterns (ULBP)}

The texture feature was extracted from LBPs as a gray-scale texture operator. It characterizes the spatial structure of the local image texture \cite{Ojala2002}. Given a central pixel in the image, a binary pattern number was computed by comparing its value with those of its neighbors. The original operator used a 3x3 window size that contains nine values. We computed the LBP features from the pixel intensities in a neighborhood.

\begin{equation}\label{eq:LBP}
LBP_{P,R}(x,y)=\bigcup_{(x',y')\in N(x,y)}h(I(x,y),I(x',y'))
\end{equation}

Where $N(x,y)$ is the vicinity around $(x,y)$, $\cup$ is the concatenation operator, $P$ is the number of neighbors, and $R$ is the radius of the neighborhood.

\begin{figure*}[h]
\begin{centering}
\begin{tabular}{|c|c|c|c|}
\hline 
Raw & ULBP(8,1)&HOG(3x3) & iHOG(3x3) \tabularnewline
\hline 
\hline 
\includegraphics[scale=0.40]{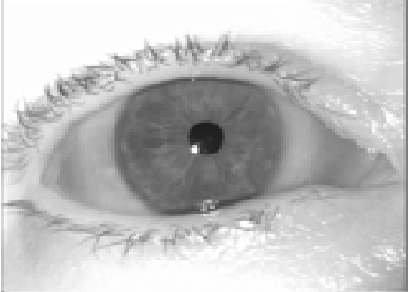} & \includegraphics[scale=0.32]{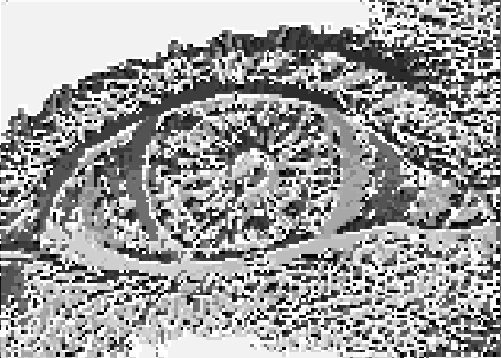}&\includegraphics[scale=0.40]{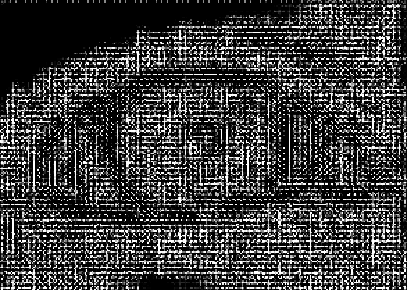}&\includegraphics[scale=0.40]{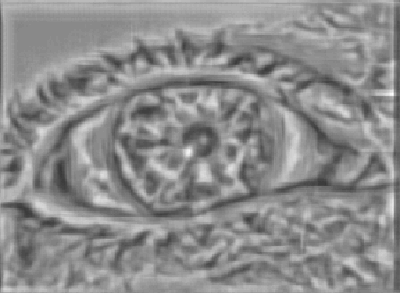} \tabularnewline
\hline 
\end{tabular}
\par\end{centering}
\caption{Features extracted from a particular right NIR periocular image without mask for intensity (raw), texture (ULBP 8,1) and shape (HOG and iHOG).}
\label{featureExtractionfigure}
\end{figure*}

\begin{figure*}[h]
\begin{centering}
\begin{tabular}{|c|c|c|c|}
\hline 
{a}&{b}&{c}&{d}\tabularnewline 
\hline 
\hline 
\includegraphics[scale=0.18]{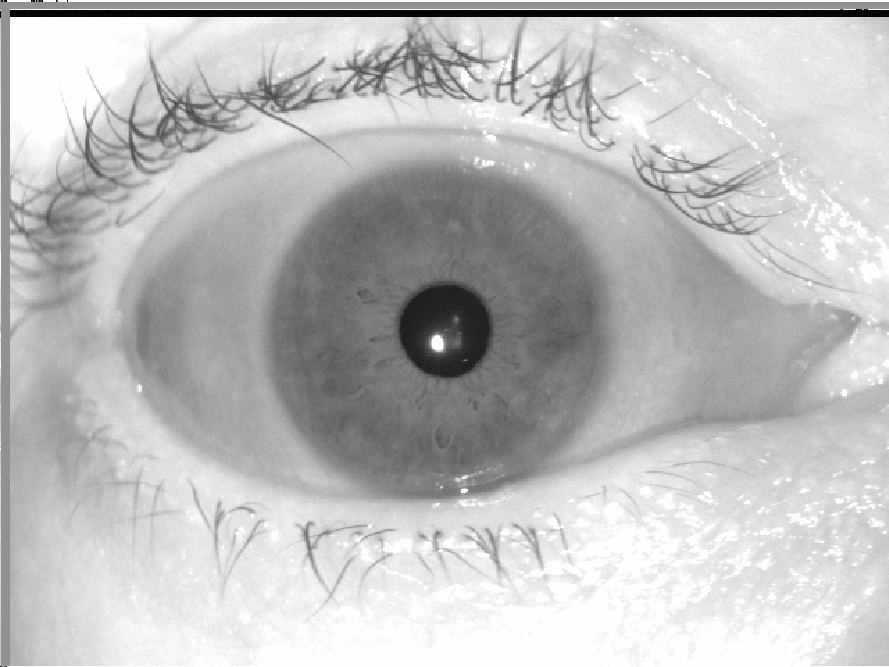}& 
\includegraphics[scale=0.18]{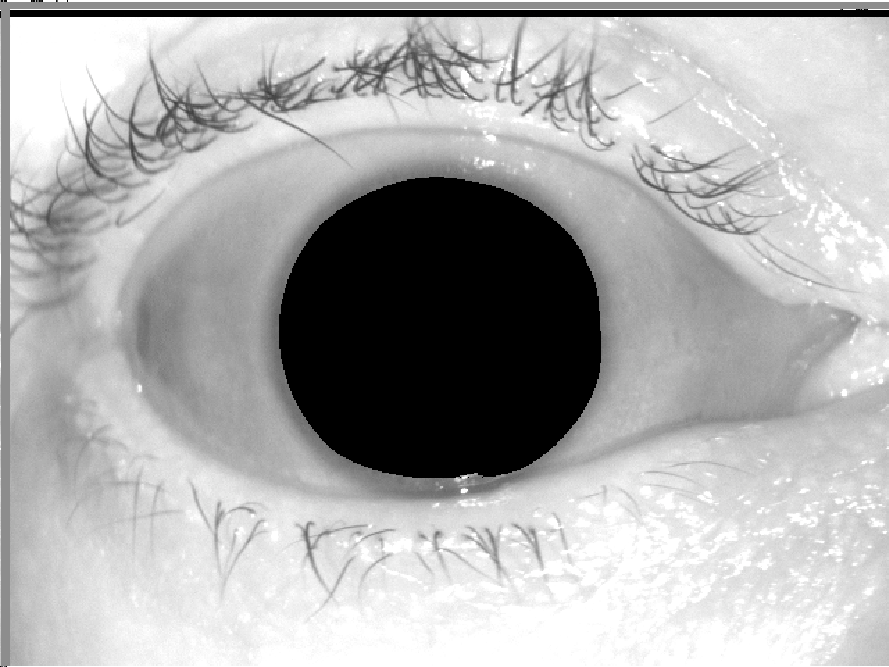}&
\includegraphics[scale=0.11]{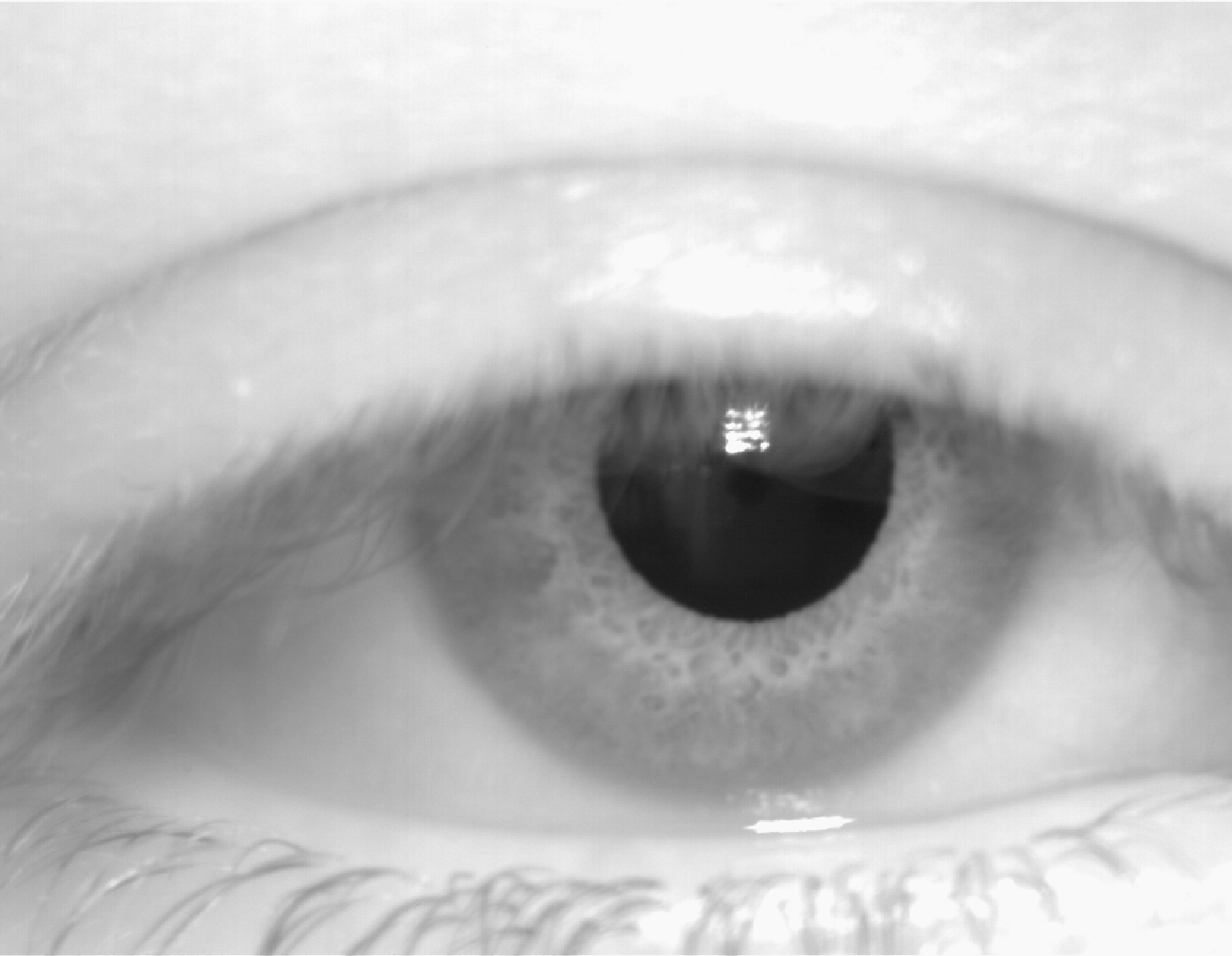}&
\includegraphics[scale=0.29]{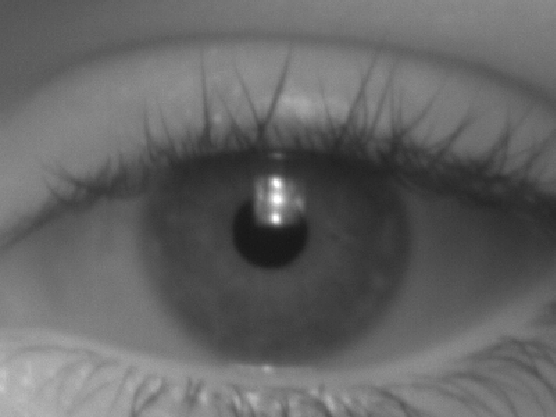} \tabularnewline  
\hline
\end{tabular}
\par\end{centering}
\caption{\label{fig:muestras} NIR Images from the right and left eyes: (a) Example of Periocular UND image, (b) Example of Periocular UND image, with the iris area covered automatically in black, (c) Example of UTIRIS images, (d) Example of Cross-Eyed 2016 images.}

\end{figure*}

Ojala et al. \cite{Ojala2002} defined a Uniform Local Binary Pattern (ULBP) to improve the traditional approach using all circular transitions $2^{8}=256$. The uniform mapping produces only 59 output labels for neighborhoods when having eight sampling points. An LBP is called uniform when its uniformity measure is of 2 transitions. For instance, pattern 01110000 shows 2 transitions so it is uniform. The texture recognition improves when using uniform patterns of (8, 1) neighborhood \cite{GuoZhangZhang2010}. This was also shown for gender classification in faces \cite{Tapia2013,Shan2012431}(See Figure \ref{featureExtractionfigure}).

\subsection{Shape - Histogram of gradients}

The shape features were extracted from the HOG and inverse HOG using three different scales: $3\times3$, $5\times5$, and $10\times10$ \cite{DalalTriggs2005, VondrickKhoslaMalisiewiczEtAl2013}. Vertical and horizontal edge maps were computed using the masks $\left[-1,0,1\right]$ and $\left[-1,0,1\right]^{T}$. We considered $v$ and $h$ to be the vertical and horizontal edge values at any pixel. They were obtained by the convolution of the edge mask with the original image respectively. The edge map was computed using $\theta=tan^{-1}\left(\frac{v}{h}\right)$, and the edge magnitude was computed as $m=\sqrt{v^{2}+h^{2}}$. The edge map was discretized at 18$^{\circ}$ intervals. Each pixel contribute its magnitude $m$ to the bin that corresponded to its edge directions $\theta$. An image is represented by $20\times N$ reals values. Where $N$ is the number of windows (See Figure \ref{featureExtractionfigure}).

\section{Databases}\label{databases}

One of the most challenging problems in gender-classification when using iris or periocular images is the lack of a dataset for the training process. Most of the databases were created with a focus on the iris/face identification problem. Therefore, databases with gender information are private and usually not available for research. The databases usually do not have enough subjects.

In this paper, we use five databases that include gender information. The first one is a subset from the GFI-UND \cite{TapiaPerezBowyer2016}. This database was captured using an LG-4000 iris sensor with full periocular images for the left and right eyes. Examples are shown in Figure 3a. 
The UND periocular Dataset is a new dataset different than used in \cite{TapiaPerezBowyer2016}. This dataset will be available to other researchers upon request. 

The second one is the same database, but with the iris masked in black for the left and right eyes.  Examples are shown in Figure 3b. 
Those two datasets are person-disjoint and have one left eye image and one right eye image for each male and female. Note that a disjoint database assures that iris images are only present in training or the testing partition, but not in both.

The left full eye dataset has a partition of 952 subjects (462 female images and 490 male images). The right full eye dataset, on the other hand, has 946 subjects (458 female images and 488 male images). Six images were removed from the dataset because they caused errors on the automatic segmentation mask. https://www.overleaf.com/project/5c1a5f3c3235f057efbfb5a9

The third database is the University of Tehran IRIS database (UTIRIS) image repository \cite{HosseiniAraabiSoltanian-Zadeh2010}. This database is the first iris biometric database registered in two distinct sessions of VIS and NIR images.

The database has 389 images obtained from 79 individuals for both right and left eyes (326 male images and 63 female images. This database is available to other researchers \footnote{https://utiris.wordpress.com/2014/03/04/university-of-tehran-iris-image-repository/}.  The people are enumerated in the same way as in NIR and VIS sessions. This database was obtained with an ISG Lightwise LW camera, and the image dimension is 1000 x 776 pixels. An example of
database is shown in Figure 3c.

The fourth database was the CROSS-EYED, Reading Cross-spectrum Iris/Periocular Database \cite{ASLC2016}. This is the benchmark dataset for the identification competition presented in BTAS 2016. This database is available to other researchers \footnote{https://sites.google.com/site/crossspectrumcompetition/cross-eyed-2016}.  It has eye images captured at the same time using  NIR and Visible (RGB) VIS sensors. The images were acquired from a distance of approximately 1.5 m. The  NIR wavelength has a single channel of information, while the VIS iris images contain three channels. The images present an indoor environment with realistic illumination conditions. They also include large variations in ethnicity and eye color as well as realistic and challenging illumination reflections. The database has 240 subjects. There are two folders for each subject: one for the NIR and the other one for VIS images. For each folder, there are eight images. In total, we have 632 male images and 328 female images (1,920 images). An example of the database is illustrated in Figure 3d. 

\subsection{UNAB-Gender Database}

As an additional contribution to this paper, we created a database in order to facilitate research in this area. A database at Universidad Andres Bello called  UNAB-Gender was created. These images provide to researchers a person-disjoint set for evaluating the gender-classification problem.

This database was also to test and compare our approach with other databases. \footnote{The UNAB-Gender is available to download in https://jtapiafarias.wordpress.com/biometric/database/}.
The images were captured using the iCAM TD-100 NIR sensor. The iCAM TD-100 uses near-infrared illumination and acquires 480x640 8-bit pixels per image. This set of iris images was obtained over five sessions with 69 female and 66 male subjects. In total, we obtained 1,412 female images and 1,356 male images. All the images are labeled with the gender information.

It is important to note that all the dataset samples from all source were represented in a balanced manner in both training and testing datasets.

\section{Experiments and Results}

In this section, we describe five experiments and seven tests designed to test the performance of the gender classification when using the iris image. An SVM and nine ensemble classifiers as shown in Tables \ref{tab-und}, and \ref{tab-class} were used.

The experiments were designed to try different features extracted from the datasets. The tests, on the other hand, were designed to test subsets taken from the databases described in section \ref{databases}.

For training purpose, we used the UND periocular dataset because has a bigger number of disjoint images. A training portion of 952 and 946 periocular UND images were used for the left and right eyes, respectively. 

All the datasets were selected using a person-disjoint method,  60\% of the males and 60\% of the females images but taking into account a k-fold cross-validation function that ensures the images are unique in each dataset and also unique when considering both datasets together. Therefore, images in the testing set are not present in the validation set.

Once the parameter selection was finalized, the selected parameterization of the method was trained on the full 60\% of training data, and we performed a single evaluation on 40\% of the test data. In each experiment, an SVM classifier with a Gaussian kernel was trained using LIBSVM implementation \cite{Chang2011}. To validate the results, we used NIR-UTIRIS, NIR-Cross-Eyed 2016 and NIR-UNAB-Gender databases to analyze how well the classifier generalizes the new data.

\subsection{Experiments}

We performed five experiments using different inputs for the SVM classifier. 
\subsubsection*{Experiment 1}: In this experiment, we use the raw pixel intensity as the input of the SVM classifier. 

\subsubsection*{Experiment 2}: In this case, the inputs to the SVM classifier are histograms from ULBP (8,1) up to ULBP (8,8). Where each histogram was build considering 59 bins.

\subsubsection*{Experiment 3}: In this experiment, the inputs to the SVM classifier are a concatenation of ULBP feature extracted from each image (from ULBP (8,1) to ULBP (8,8) in the radii ULBP database) The resulting length is 472 ($59 features\times8 radii = 472 bins$) for each periocular image. 
\subsubsection*{Experiment 4}: For this experiment, the inputs to the SVM classifier are HOG and IHOG features obtained using block sizes of $3\times3$, $5\times5$, and $10\times10$, respectively. 
In the last three experiments (from 2 to 4) we use a histogram of features. That allows us to reduce the number of features used in the SVM classifier. This reduces considerably the timing of the process which is a factor in most real-time applications. 
 
\subsubsection*{Experiment 5}: For Experiment (5), we concatenated the information from the histogram of intensity, histogram of texture and histogram of shape (Pixel Values + ULBP(8,1 to 8,8) + HOG ($3\times3$, $5\times5$, $10\times10$)).

\subsection {Tests for validation}
We defined seven tests for validating the correct classification rate. For each test, we use a subset taken from the databases described previously.

\subsubsection*{Test 1}: In this test, we have considered the correct classification rate using the full UND periocular NIR images (Non occluded), as illustrated in Figure 3a. 

\subsubsection*{Test 2}: In this test, we used the iris artificially masked (occluded condition) from UND periocular NIR images, as illustrated in Figure 3b. The segmentation stage automatically finds the iris and the pupil. It also masks in black the highlight, occlusion and the iris from the periocular image. The normalization, encoding, and comparison stages are not needed in this approach.

\subsubsection*{Test 3}: In this test, we added a new dataset named UTIRIS. The previous two tests, this dataset was not part of the training set. Therefore, the results allow us to understand how well the previous classifier (Test 1 and Test 2) generalized to the new data. In this test, we considered the correct classification rate using an SVM model that was trained with the UND periocular database (Non-occluded and Occluded) (see Figure 3c).

\subsubsection*{Test 4}: For additional validation, we use the Cross-Eyed 2016 database. This dataset, like the previous one, was not part of the training set. In this test, we considered the correct classification rate using an SVM model trained with the UND periocular database (Non-occluded and Occluded) (See Figure 3d).

\subsubsection*{Test 5}: In this test, we use the new database created for this paper, the UNAB-Gender NIR images (See Figure 3c). The database was not part of the training set.

\subsubsection*{Test 6}: In the previous test (from 1 to 5), we have used the UND Periocular database as a training set. In this case, in particular, we used images from the following databases: UND Periocular+UTIRIS+Cross-Eyed. This combined database was used for training and testing. The use of three features (Intensity, shape, and texture) combined at different scales performed better than using only one feature on a single scale.

In all the experiments, we estimated the correct classification rate separately for the left and right eyes. Since some of the capture systems worked with a single periocular image. For systems capturing both irises, we can use either of them to validate the identification.
 
The results for the correct classification rate obtained for Tests 1 to 6 are reported in Table \ref{tab-uti}, Table\ref{tab-ce}, Table\ref{tab-unab}, and Table\ref{tab-all}, respectively. In these tables we report results when using SVM classifier. 

Note that we are using different databases where the images may not present exactly the same conditions. Even images that were acquired at the same time (i.e left and right eye) may present different illumination conditions due to sensor position. Specular highlights may also be present in different parts of the image. The inclination of the head, the eyelid occlusion, and the type of camera used may also modify the image. 

\section{Discussion}

In this section we present the results for the six tests and five experiments (section \ref{genderresults}). In Section \ref{Features} we discuss the most relevant information for gender
classification when using periocular images. A statistical ANOVA test to validate our results is shown in section \ref{ANOVA}.

\subsection{Gender Classifications results}\label{genderresults}
Table \ref{tab-und} shows the correct classification rate achieved from periocular images (Non  occluded  and Occluded) when the images are trained  and  tested  using  UND  Periocular database (test 1 and 2, experiments 1 to 5). The periocular UND database presents a similar balance between male and female images. For the Non occluded periocular images of the left eyes, the best result was achieved by the fusion of HOG $5\times5$ with 82.83 +/- 0.5\%. For the right eyes, the best result was achieved by the concatenation of HOG $10\times10$ with 85.74 +/- 0.4\%. 
For the occluded or masked periocular images of the left eyes, the best result was HOG $10\times10$ with 80.00 +/- 0.5\%. For the right eyes, the best result was also achieved by the concatenation of ULBP(8,1) to (8,8) with 84.33 +/- 0.6\%. 

Based on these results, we can state that there is no significant influence of iris information on correct classification rate when using periocular images. We observed a variation of approximately 4\% between the best results when using Non occluded and Occluded images. Table \ref{tab-class}, also demonstrate this finding when using the nine different classifiers applied to the best experiments and tests from Table \ref{tab-und}.

\begin{table}[h]
\scriptsize
\centering
\caption{Correct classification rates for the left and right periocular images trained and tested using \textbf{UND Periocular} considering the comparison between the Non occluded and Occluded iris images.} \label{tab-und}
\begin{tabular}{|c|c|c|c|c|}
\hline
\multirow{3}{*}{Method} & \multicolumn{2}{c|}{Test 1}                   & \multicolumn{2}{c|}{Test 2}                   \\ \cline{2-5} 
                        & Left eye              & Right eye             & Left eye              & Right eye             \\ \cline{2-5} 
                        & Non occluded          & Non occluded          & Occluded              & Occluded              \\ \hline 
\multicolumn{5}{|c|}{\textit{Experiment 1}} \tabularnewline
\hline
{Raw}  & {53.57+/-0.7} & {51.57 +/- 0.7} & {54.57 +/- 0.7}& { 53.57+/- 0.7}                                                                                     \tabularnewline
\hline
\multicolumn{5}{|c|}{\textit{Experiment 2}}                                                                                                                            \tabularnewline
\hline
ULBP (8,1)                                                    & 73.68 +/- 0.6          & 78.83+/- 0.6           & 75.78 +/- 0.8          & 75.66 +/- 0.8          \tabularnewline
ULBP (8,2)                                                    & 74.73 +/- 0.6          & 73.80 +/- 0.7          & 76.31 +/- 0.8          & 75.92 +/- 0.8          \tabularnewline 
ULBP (8,3)                                                    & 74.73 +/- 0.7          & 74.07 +/- 0.7          & 75.78 +/- 0.7          & 74.60 +/- 0.6          \tabularnewline
ULBP (8,4)                                                    & 74.73+/- 0.6           & 74.60 +/- 0.7           & 79.47 +/- 0.7          & 74.60 +/- 0.6          \tabularnewline
ULBP (8,5)                                                    & 76.31 +/- 0.6          & 78.83 +/- 0.6          & 78.42 +/- 0.7          & 78.30 +/- 0.6          \tabularnewline 
ULBP (8,6)                                                    & 77.89 +/- 0.6          & 74.60 +/- 0.6          & 76.31+/- 0.7           & 77.24,+/- 0.5          \tabularnewline 
ULBP (8,7)                                                    & 74.21 +/- 0.6          & 73.80 +/- 0.6          & 75.52 +/- 0.7          & 77.24,+/- 0.5          \tabularnewline
ULBP (8,8)                                                    & 74.73 +/- 0.6          & 77.24+/- 0.6           & 76.31+/- 0.7           & 75.66 +/- 0.6          \tabularnewline
\hline
\multicolumn{5}{|c|}{\textit{Experiment 3}}                                                                                                                             \tabularnewline
\hline
\begin{tabular}[c]{@{}c@{}}ULBP\\ (8,1- 8,8)\end{tabular} & 78.68 +/- 0.6          & 80.42 +/- 0.6          & 78.82 +/- 0.7          & \textbf{84.33 +/- 0.6} \tabularnewline
\hline
\multicolumn{5}{|c|}{\textit{Experiment 4}}                                                                                                                           \tabularnewline
\hline
\begin{tabular}[c]{@{}c@{}}HOG\\ (3x3)\end{tabular}           & 78.42 +/- 0.6          & 82.01 +/- 0.6          & 76.57 +/- 0.7          & 79.01 +/- 0.8          \tabularnewline
\begin{tabular}[c]{@{}c@{}}HOG\\ (5x5)\end{tabular}           & \textbf{82.83 +/- 0.5} & 83.06+/- 0.5           & 74.73 +/- 0.4          & 78.04 +/- 0.7          \tabularnewline 
\begin{tabular}[c]{@{}c@{}}HOG\\ (10x10)\end{tabular}         & 80.52 +/- 0.6          & \textbf{85.74 +/- 0.4} & \textbf{80.00 +/- 0.5} & 83.59 +/- 0.6         \tabularnewline
\hline
\multicolumn{5}{|c|}{\textit{Experiment 5}}                                                                                                                           \tabularnewline
\hline
\begin{tabular}[c]{@{}c@{}}Fusion\\ \end{tabular}          &  81.84       &   83.60        &  82.63         &  82.01         \tabularnewline
\hline
\end{tabular}
\end{table}

In order to show that the results are not dependent the SVM classifier we included new experiments using other classifiers applied to the best results of Table \ref{tab-und}.
We used nine 'ensemble classifiers' with the periocular UND images: AdaboostM1, LogitBoost, GentleBoost, RobustBoost, LPBoost, TotalBoost, RUSBoost with 'Tree learners' classifiers and learning rate of 0.1, Random Forest classifier (RF) with 900 'Trees'. The results are presented in Table \ref{tab-class}.

\begin{table}[h]
\scriptsize
\centering
\caption{Correct classification rates with method \textbf{HOG/LBP} for nine ensemble classifier applied to the best results of Table \ref{tab-und}. B represents: Boost.} \label{tab-class}
\begin{tabular}{|c|c|c|c|c|}
\hline
\multirow{2}{*}{\textbf{Classifier}} & \textbf{Left eye (\%)} & \textbf{Right eye (\%)} & \textbf{Left eye (\%)} & \textbf{Right eye (\%)}                   \\ \cline{2-5} 
                                     & \textbf{Non occluded}  & \textbf{Non occluded}   & \textbf{Occluded}      & \textbf{Occluded}              \\ \hline 
\textbf{SVM}                         & 82.83 +/- 0.5         & 85.74 +/- 0.4          & 80.00 +/-0.5          & 84.33 +/- 0.6           \tabularnewline
\hline
\textbf{Bag}                         & 79.52 +/- 0.3         & 84.66 +/- 0.3           & 81.33 +/- 0.3         & 82.33 +/- 0.6           \tabularnewline
\hline 
\textbf{Ada-B}                        & 78.66 +/- 0.6         & 84.33 +/- 0.6          & 81.33 +/- 0.6         & 81.66 +/- 0.6           \tabularnewline
\hline
\textbf{Logit-B}                  & 80.33 +/- 0.3         & 79.66 +/- 0.6          & 80.66 +/- 0.3         & 83.33 +/- 0.3           \tabularnewline
\hline  
\textbf{Gentle-B}                 & 81.00 +/- 0.3         & 79.66 +/- 0.3          & 79.66 +/- 0.6         & 80.66 +/- 0.6   \tabularnewline
\hline  
\textbf{Robust-B}                 & 79.66 +/- 0.3         & 82.33 +/- 0.6          & 80.66 +/- 0.3         & 81.33 +/- 0.3           \tabularnewline
\hline  
\textbf{LP-B}                     & 81.66 +/- 0.6         & 83.33 +/ 0.3           & 81.00 +/- 0.3         & 82.33 +/- 0.6           \tabularnewline
\hline  
\textbf{Total-B}                  & 79.66 +/- 0.6         & 81.33 +/- 0.6          & 79.66 +/- 0.6         & 80.33 +/- 0.3           \tabularnewline
\hline  
\textbf{Rus-B}                    & 78.66 +/- 0.3         & 81.66 +/- 0.3          & 78.66 +/- 0.6         & 80.66 +/- 0.3          \tabularnewline
\hline  
\textbf{RF GDI}                   & 81.33 +/- 0.6         & 84.66 +/- 0.5          & 81.66 +/- 0.6         & 84.00 +/- 0.3           \tabularnewline
\hline
\end{tabular}
\end{table}

Table \ref{tab-uti} (test 3, experiments 1 to 5), Table \ref{tab-ce} (test 3, experiments 1 to 5), Table \ref{tab-unab} (test 4, experiments 1 to 5),  and Table \ref{tab-all} (test 5, experiments 1 to 5), also show the correct classification rate. The results on the Table indicate the performance of the classifier. 

\begin{table}[h]
\scriptsize
\centering
\caption{Correct classification rates for the left and right periocular images trained with \textbf{UND database} (Non occluded and Occluded) and tested with \textbf{UTIRIS NIR}.} \label{tab-uti}
\begin{tabular}{|c|c|c|c|c|}
\hline
\multirow{3}{*}{Method} & \multicolumn{2}{c|}{Test 1}                   & \multicolumn{2}{c|}{Test 2}                   \\ \cline{2-5} 
                        & Left eye              & Right eye             & Left eye              & Right eye             \\ \cline{2-5} 
                        & Non occluded          & Non occluded          & Occluded              & Occluded              \\ \hline 
\multicolumn{5}{|c|}{\textit{Experiment 1}} \tabularnewline
\hline
Raw & {55.60+/-0.6} & {57.53 +/- 0.5} & {55.90 +/- 0.8}  & {58.00+/- 0.3}                                                                                     \tabularnewline
\hline
\multicolumn{5}{|c|}{\textit{Experiment 2}}                                                                                                                            \tabularnewline
\hline
ULBP (8,1)              & 80.55 +/- 1.0        & 83,80+/- 1.0         & 85.09 +/- 0.8        & 85.35 +/- 0.8        \tabularnewline
ULBP (8,2)              & 83.55 +/- 0.9        & 79.18 +/- 0.9        & 85.09 +/- 0.8        & 66.32+/- 0.9         \tabularnewline
ULBP (8,3)              & 83.80 +/- 0.8        & 80.46 +/- 0.8        & 84.32 +/- 0.7        & 70.44 +/- 0.7        \tabularnewline 
ULBP (8,4)              & 80.98 +/- 0.9        & 68.64 +/- 0.8        & 76.91+/- 0.7         & 82.78 +/- 0.6        \tabularnewline 
ULBP (8,5)              & 83.03 +/- 0.8        & 66.84 +/- 0.8        & 79.18 +/- 0.7        & 84.83 +/- 0.6        \tabularnewline
ULBP (8,6)              & 82.01 +/- 0.8        & \textbf{85.35 +/- 0.6}        & 82.26+/- 0.7         & 82.26 +/- 0.6        \tabularnewline
ULBP (8,7)              & 81.75 +/- 0.8        & 81.32 +/- 0.7        & 81.49 +/- 0.7        & 77.63 +/- 0.6        \tabularnewline
ULBP (8,8)              & 81.75 +/- 0.6        & 77.12+/- 0.7         & 53.21+/- 0.7         & 74.04,+/- 0.6        \tabularnewline
\hline
\multicolumn{5}{|c|}{\textit{Experiment 3}}                                                                                                                             \tabularnewline
\hline
\begin{tabular}[c]{@{}c@{}}ULBP\\ (8,1- 8,8)\end{tabular} & 84.32+/- 0.5         & 84.58 +/- 0.6        & 84.32 +/- 0.7        & 64.27+/- 0.6 \tabularnewline
\hline
\multicolumn{5}{|c|}{\textit{Experiment 4}}                                                                                                                           \tabularnewline
\hline
\begin{tabular}[c]{@{}c@{}}HOG\\ (3x3)\end{tabular}           & 71.12 +/- 0.6        & 72.24 +/- 0.6        & 69.92 +/- 0.7        & 58.61 +/- 0.6          \tabularnewline
\begin{tabular}[c]{@{}c@{}}HOG\\ (5x5)\end{tabular}           & 57.58 +/- 0.6        & 65.04+/- 0.6         & 41.13 +/- 0.7        & 47.59+/- 0.6          \tabularnewline 
\begin{tabular}[c]{@{}c@{}}HOG\\ (10x10)\end{tabular}         & 52.19 +/- 0.6        & 60.33 +/- 0.5        & 56.22 +/- 0.7        & 57.22+/- 0.6         \tabularnewline
\hline
\multicolumn{5}{|c|}{\textit{Experiment 5}}                                                                                                                           \tabularnewline
\hline
\begin{tabular}[c]{@{}c@{}}Fusion\\ \end{tabular}         & \textbf{85.60}                 & 82.26                 & \textbf{85.26}                 & \textbf{86.89}         \tabularnewline
\hline
\end{tabular}
\end{table}

\begin{table}[h]
\scriptsize
\centering
\caption{Correct classification rates for the left and right periocular images trained with \textbf{UND database} (Non occluded and Occluded) and tested with \textbf{Cross-Eye-2016 NIR}.} \label{tab-ce}
\begin{tabular}{|c|c|c|c|c|}
\hline
\multirow{3}{*}{Method} & \multicolumn{2}{c|}{Test 1}                   & \multicolumn{2}{c|}{Test 2}                   \\ \cline{2-5} 
                        & Left eye              & Right eye             & Left eye              & Right eye             \\ \cline{2-5} 
                        & Non occluded          & Non occluded          & Occluded              & Occluded              \\ \hline 
\multicolumn{5}{|c|}{\textit{Experiment 1}} \tabularnewline
\hline
Raw                 & 60.61+/-0.3            & 59.30 +/- 0.5          & 59.33 +/- 0.6         & 59.00+/- 0.6        \tabularnewline
\hline
\multicolumn{5}{|c|}{\textit{Experiment 2}}                                                                         \tabularnewline
\hline
ULBP (8,1)              & 68.96 +/- 0.9          & 69.38+/- 0.8          & 72.40 +/- 0.8         & 74.69 +/- 0.8          \tabularnewline
ULBP (8,2)              & 68.94 +/- 0.7          & 74.48 +/- 0.7         & 70.63+/- 0.8          & 66.98+/- 0.9           \tabularnewline 
ULBP (8,3)              & 72.40 +/- 0.6          & 79.90 +/- 0.6         & 78.85 +/- 0.7         & 74.90 +/- 0.7          \tabularnewline
ULBP (8,4)              & 77.19 +/- 0.9          & 52.08 +/- 0.9         & 77.40 +/- 0.7         & 71.88 +/- 0.6          \tabularnewline
ULBP (8,5)              & 72.67 +/- 0.6          & 68.65 +/- 0.6         & \textbf{80.21+/- 0.9} & \textbf{72.29 +/- 0.6} \tabularnewline
ULBP (8,6)              & 77.50 +/- 0.6          & 77.29 +/- 0.6         & 75.00+/- 0.7          & 74.58 +/- 0.6          \tabularnewline
ULBP (8,7)              & 77.50 +/- 0.7          & 78.23 +/- 0.7         & 79.48 +/- 0.5         & 76.35 +/- 0.6          \tabularnewline
ULBP (8,8)              & \textbf{78.85 +/- 0.6} & \textbf{79.90 +/- 0.9} & 77.92+/- 0.7          & 75.10 +/- 0.6          \tabularnewline
\hline
\multicolumn{5}{|c|}{\textit{Experiment 3}}                                                                              \tabularnewline
\hline
\begin{tabular}[c]{@{}c@{}}ULBP\\ (8,1- 8,8)\end{tabular} & 75.10+/- 0.6           & 76.88 +/- 0.6         & 76.04 +/- 0.7         & 73.96+/- 0.6 \tabularnewline
\hline
\multicolumn{5}{|c|}{\textit{Experiment 4}}                                                                              \tabularnewline
\hline
\begin{tabular}[c]{@{}c@{}}HOG\\ (3x3)\end{tabular}           & 73.54 +/- 0.9          & 71.56 +/- 0.6         & 75.10 +/- 0.7         & 65.10 +/- 0.6          \tabularnewline
\begin{tabular}[c]{@{}c@{}}HOG\\ (5x5)\end{tabular}           & 67.08 +/- 0.6          & 74.79 +/- 0.6         & 57.92 +/- 0.7         & 51.88+/- 0.6          \tabularnewline 
\begin{tabular}[c]{@{}c@{}}HOG\\ (10x10)\end{tabular}         & 74.58 +/- 0.6          & 63.23 +/- 0.5         & 55.21 +/- 0.9         & 57.29+/- 0.6         \tabularnewline
\hline
\multicolumn{5}{|c|}{\textit{Experiment 5}}                                                                                                                           \tabularnewline
\hline
\begin{tabular}[c]{@{}c@{}}Fusion\\ \end{tabular}         & 77.71                   & 77.92                  & 70.73                  & 71.77         \tabularnewline
\hline
\end{tabular}
\end{table}

\begin{table}[h]
\scriptsize
\centering
\caption{Correct classification rates for the left and right periocular images trained with \textbf{UND database} (Non occluded and Occluded) and tested with \textbf{UNAB-Gender}.} \label{tab-unab}
\begin{tabular}{|c|c|c|c|c|}
\hline
\multirow{3}{*}{Method} & \multicolumn{2}{c|}{Test 1}                   & \multicolumn{2}{c|}{Test 2}                   \\ \cline{2-5} 
                        & Left eye              & Right eye             & Left eye              & Right eye             \\ \cline{2-5} 
                        & Non occluded          & Non occluded          & Occluded              & Occluded              \\ \hline 
\multicolumn{5}{|c|}{\textit{Experiment 1}} \tabularnewline
\hline
Raw                & 54.61+/-0.66            &53.30 +/- 0.3          & 53.66 +/- 0.6          & 55.00+/- 0.6        \tabularnewline
\hline
\multicolumn{5}{|c|}{\textit{Experiment 2}}                                                                         \tabularnewline
\hline
ULBP (8,1)              & 71.66 +/- 0.9          & 72.33+/- 0.8          & 72.66 +/- 0.8         & 73.66 +/- 0.8          \tabularnewline
ULBP (8,2)              & 71.66 +/- 0.7          & 74.48 +/- 0.7         & 73.66+/- 0.8          & 72.33+/- 0.9           \tabularnewline
ULBP (8,3)              & 72.40 +/- 0.6          & 76.90 +/- 0.6         & 78.33 +/- 0.8         & 74.33 +/- 0.5          \tabularnewline
ULBP (8,4)              & 77.33 +/- 0.9          & 72.33 +/- 0.9         & 77.33 +/- 0.3         & 72.66 +/- 0.6          \tabularnewline
ULBP (8,5)              & 74.66 +/- 0.3          & 78.65 +/- 0.3         & 78.21+/- 0.6          & 76.66 +/- 0.6        \tabularnewline
ULBP (8,6)              & 77.33 +/- 0.6          & 73.33 +/- 0.6         & 74.66+/- 0.7          & 76.66 +/- 0.6          \tabularnewline
ULBP (8,7)              & 77.33 +/- 0.7          & 75.66 +/- 0.7         & 76.33 +/- 0.5         & 76.33 +/- 0.3          \tabularnewline
ULBP (8,8)              & 76.66 +/- 0.4         & 80.33 +/- 0.3          & 75.33+/- 0.7          & 76.33 +/- 0.3          \tabularnewline
\hline
\multicolumn{5}{|c|}{\textit{Experiment 3}}                                                                              \tabularnewline
\hline
\begin{tabular}[c]{@{}c@{}}ULBP\\ (8,1- 8,8)\end{tabular} & 79.66+/- 0.3           & 78.66 +/- 0.6         & 76.33 +/- 0.7         & 83.33+/- 0.6 \tabularnewline
\hline
\multicolumn{5}{|c|}{\textit{Experiment 4}}                                                                              \tabularnewline
\hline
\begin{tabular}[c]{@{}c@{}}HOG\\ (3x3)\end{tabular}           & 77.33 +/- 0.6          & 80.66 +/- 0.6         & 75.33 +/- 0.6         & 75.66 +/- 0.6          \tabularnewline
\begin{tabular}[c]{@{}c@{}}HOG\\ (5x5)\end{tabular}           & 81.33 +/- 0.3          & 80.33 +/- 0.6         & \textbf{79.33 +/- 0.6}         & \textbf{81.66+/- 0.6}          \tabularnewline 
\begin{tabular}[c]{@{}c@{}}HOG\\ (10x10)\end{tabular}         & \textbf{81.66 +/- 0.6}          & \textbf{82.33 +/- 0.5}         & 77.66 +/- 0.6         & 80.66+/- 0.6         \tabularnewline
\hline
\multicolumn{5}{|c|}{\textit{Experiment 5}}                                                                                                                           \tabularnewline
\hline
\begin{tabular}[c]{@{}c@{}}Fusion\\ \end{tabular}        & 81.0                   & 79.6                  & 80.3                  & 81.6        \tabularnewline
\hline
\end{tabular}
\end{table}

In Table \ref{tab-all} we show the results for test 6 with experiments from 1 to 5. 
This test shows results when the training and testing datasets are obtained by combining the Periocular UND, Periocular UTIRIS, and Periocular Cross-Eyed databases. 
The resulting training dataset has 1,265 images and the Test set has 1,036 images. The results of the correct classification rate are presented in Table \ref{tab-all}. The ULBP 8,1 to 8,8 feature extraction achieved the best results for the left and right eyes. This shows that when combining the texture features with different scales (radii) provides better results than using only one scale.

\begin{table}[h]
\scriptsize
\centering
\caption{Correct classification rates for training the left and right periocular images with a training set of UND+UTIRIS+Cross-Eyed databases and testing with a test set of UND+UTIRIS+Cross-Eyed. TNR: True Negative Rate. TPR: True Positive Rate, \textbf{MCC: Matthews Correlation Coefficient}.} \label{tab-all}
\begin{tabular}{|c|p{0.45cm}|p{0.45cm}|p{0.45cm}|p{0.45cm}|p{0.45cm}|p{0.45cm}|p{0.45cm}|p{0.45cm}|}
\hline
\multirow{3}{*}{Method} & \multicolumn{8}{c|}{Test 6}                   \\ \cline{2-9} 
                        & \multicolumn{4}{c|}{Left eye (\%)}                  & \multicolumn{4}{c|}{Right eye (\%)}             \\ \cline{2-9} 
                        & \textbf{CCR} & \textbf{TPR}   & \textbf{TNR} & {\textbf{MCC}}  & \textbf{CCR} & \textbf{TPR} & \textbf{TNR} & \textbf{MCC}              \\ \hline 
\multicolumn{9}{|c|}{\textit{Experiment 1}} \tabularnewline
\hline
{Raw}           & {65.33}       & {67.33}     & {64.33}  & {{0.44}}  & {64.33} & {64.33} & {62.33}   & {{0.45}}        \tabularnewline
\hline
\multicolumn{9}{|c|}{\textit{Experiment 2}}                                                                         \tabularnewline
\hline
ULBP(8,1)             & 81.46             & 81.11          & 63.87      & {\textbf{0.46}}        & 82.79             & 87.43          & 76.40       & {\textbf{0.62}}       \tabularnewline
ULBP(8,2)              & 78.47             & 82.11          & 64.43       & {\textbf{0.51}}       & 80.56             & 85.54          & 72.75       & {\textbf{0.57}}       \tabularnewline
ULBP(8,3)              & 76.25             & 80.97          & 62.75       & {\textbf{0.47}}       & 76.02             & 82.38          & 67.13      & {\textbf{0.47}}        \tabularnewline
ULBP(8,4)             & 77.41             & 82.39          & 66.11       & {\textbf{0.50}}       & 81.72             & 86.66          & 75.00      & {\textbf{0.60}}        \tabularnewline
ULBP(8,5)             & 79.15             & 84.19          & 70.03       & {\textbf{0.54}}       & 81.43             & 86.05          & 73.60      & {\textbf{0.59}}        \tabularnewline
ULBP(8,6)            & 76.35             & 82.20          & 66.39      & {\textbf{0.48}}        & 79.98             & 85.10          & 71.91       & {\textbf{0.56}}       \tabularnewline
ULBP(8,7)             & 77.70             & 82.46          & 66.11      & {\textbf{0.50}}        & 78.92             & 82.39          & 64.89       & {\textbf{0.52}}       \tabularnewline
ULBP(8,8)              & 77.22             & 82.43          & 66.39      & {\textbf{0.49}}        & 78.05             & 82.45          & 65.73       & {\textbf{0.51}}       \tabularnewline
\hline
\multicolumn{9}{|c|}{\textit{Experiment 3}}                                                                              \tabularnewline
\hline
\begin{tabular}[c]{@{}c@{}}ULBP\\ (8,1- 8,8)\end{tabular} & 83.49    & 84.59 & 70.35  & \textbf{0.56}   & {83.56}    & {87.46} & {76.12}   & {{0.64}} \tabularnewline
\hline
\multicolumn{9}{|c|}{\textit{Experiment 4}}                                                                              \tabularnewline
\hline
\begin{tabular}[c]{@{}c@{}}HOG\\ (3x3)\end{tabular}           & 79.63             & 84.11          & 69.47        & {\textbf{0.55}}      & 82.79             & 86.76          & 74.72        & {\textbf{0.62}}          \tabularnewline
\begin{tabular}[c]{@{}c@{}}HOG\\ (5x5)\end{tabular}           & 82.14             & 84.59          & 69.11      & {\textbf{0.60}}        & 82.59             & 85.27          & 70.79       & {\textbf{0.61}}          \tabularnewline 
\begin{tabular}[c]{@{}c@{}}HOG\\ (10x10)\end{tabular}         & 80.41             & 84.60          & 68.35       & {\textbf{0.63}}       & 81.72             & 84.68          & 69.66        & {\textbf{0.59}}         \tabularnewline
\hline
\multicolumn{9}{|c|}{\textit{Experiment 5}}                                                                                                                           \tabularnewline
\hline
\begin{tabular}[c]{@{}c@{}}Fusion\\ \end{tabular}        & 79.79             & 80.63          & 79.38      & {\textbf{0.52}}        & 80.33             & 92.70          & 77.02       & {\textbf{0.57}}        \tabularnewline
\hline
\end{tabular}
\end{table}

When a dataset is unbalanced (i.e the number of samples in one class is higher than in the other class - which happens with UTIRIS and Cross-Eyed 2016 datasets), the evaluated correct classification rate of a classifier is not representative of the true performance of the classifier. For binary classification, such as problems like gender classification, we can derive two indicators: True Positive Rate(TPR) and True Negative Rate (TNR). Those indicators are commonly used to evaluate binary classifiers. TPR can be expressed as $TP / (TP+FN)$, and TNR as $TN / (TN+FP)$. 

\begin{figure*}[h]
\minipage{0.20\textwidth}
  \includegraphics[width=\linewidth]{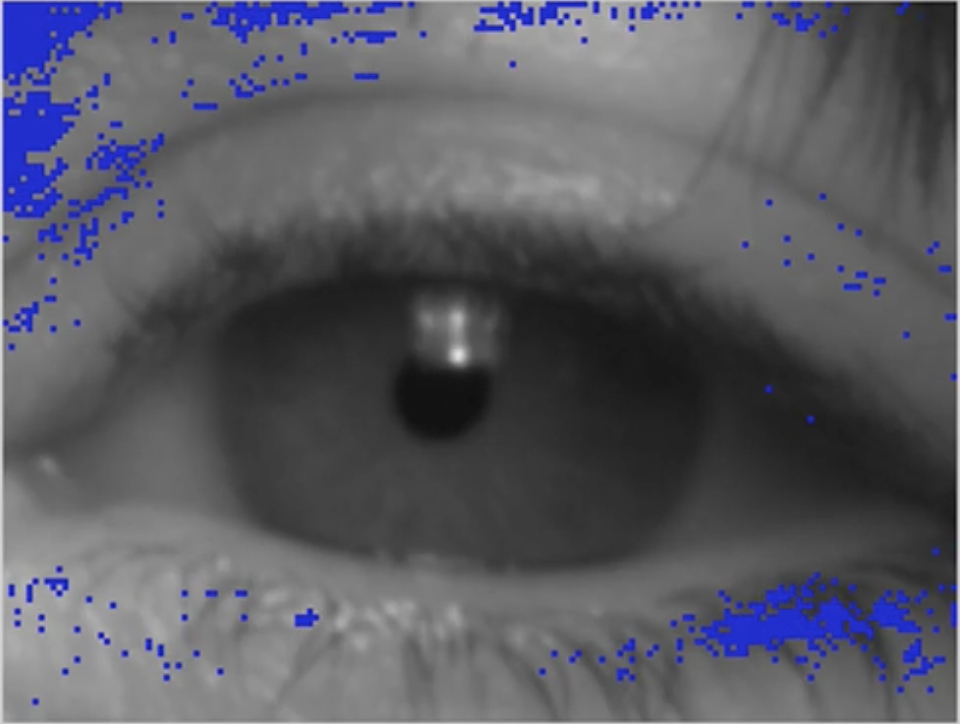}
\endminipage\hfill
\minipage{0.20\textwidth}
  \includegraphics[width=\linewidth]{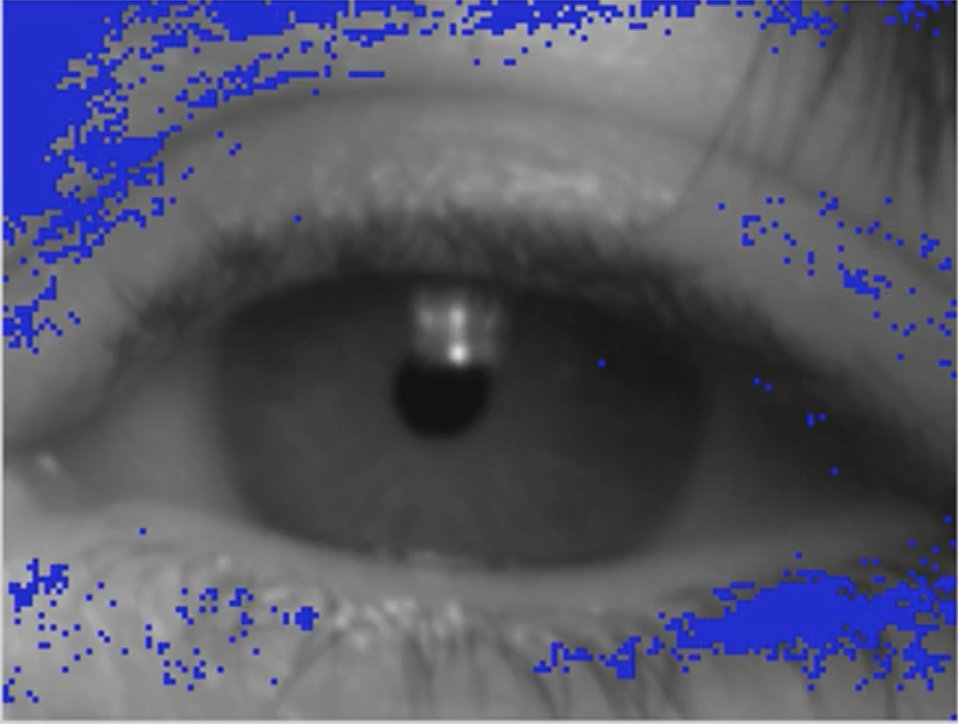}
\endminipage\hfill
\minipage{0.20\textwidth}%
  \includegraphics[width=\linewidth]{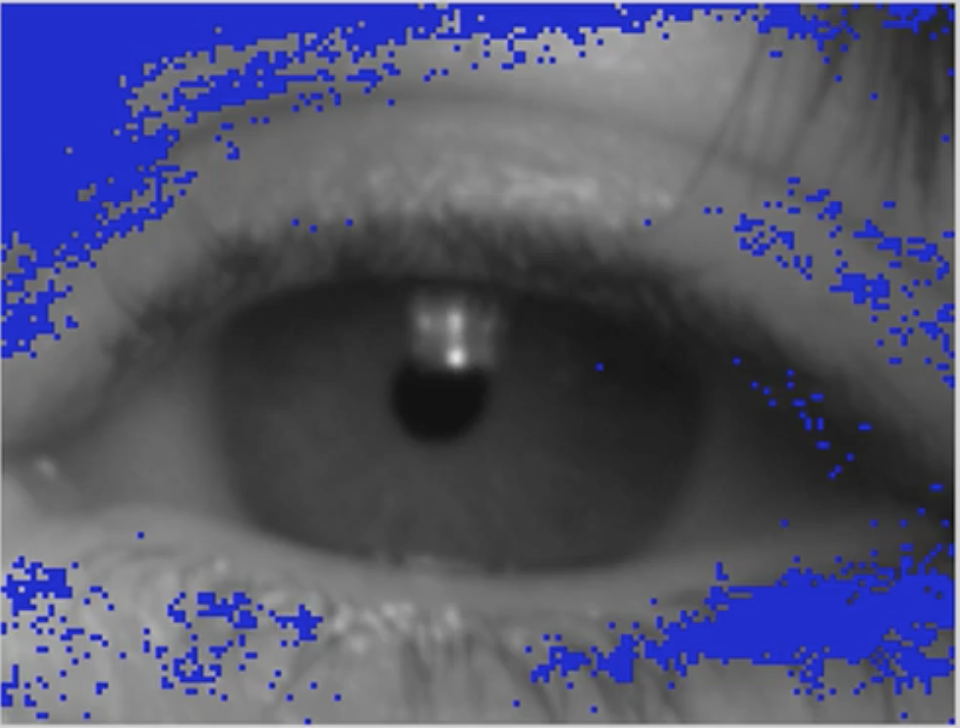}
\endminipage
\minipage{0.20\textwidth}
  \includegraphics[width=\linewidth]{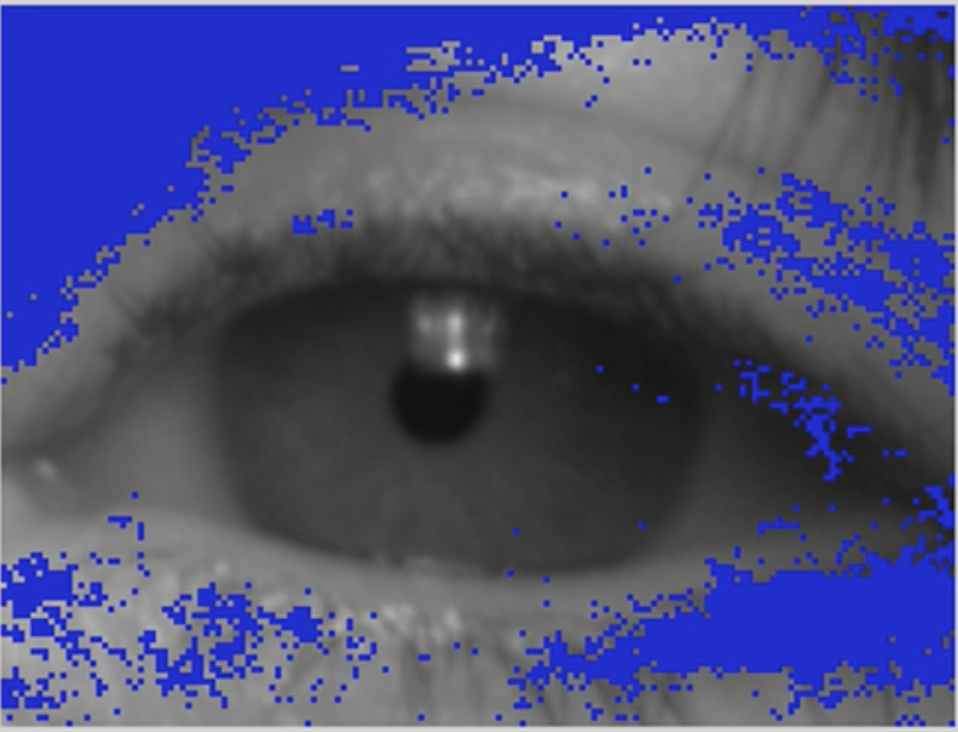}
\endminipage\hfill
\minipage{0.20\textwidth}
  \includegraphics[width=\linewidth]{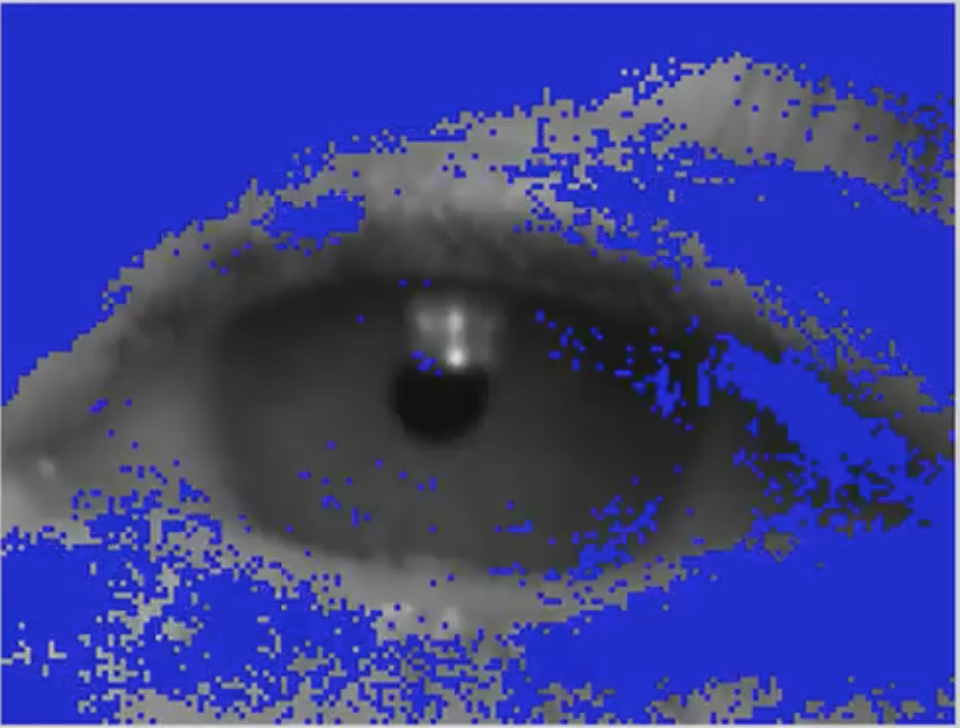}
\endminipage\hfill
\caption{Blue pixels represent incremental increase of feature selected from a random female image. From left to right: 1,000; 2,000; 3,000; 5000; 10,000 features.}
\label{sietexg}
\end{figure*}

In general, the True Negative Rate refers to the correct classification rate on the class negative (male), and the True Positive Rate refers to the correct  classification rate on the class positive (female).  We can conclude that the male prediction is more robust. This assumption is based on the high value of TPR and the low level of TNR.

{Also in {Table \ref{tableMCC} the Matthews Correlation Coefficient (MCC) \cite{boughorbel2017optimal} of the best results reached for all of the experiments from Table \ref{tab-und} to Table \ref{tab-all}. MCC takes into account true and false positives and negatives. It is a balanced measure which can be used even if the classes are of very different sizes. 
The MCC is a correlation coefficient between the observed and predicted binary classifications. MCC values returns a value between $-1$ and $+1$.  A coefficient of $+1$ represents a perfect prediction, 0 no better than random prediction and $-1$ indicates total disagreement between prediction and observation.
}}

\begin{table}[h]
\scriptsize
\centering
\caption{Summary of the best results from table \ref{tab-und} to table \ref{tab-all}. The third Column shows a MCC for the each experiment.} \label{tableMCC}
\begin{tabular}{|c|c|c|}
\hline
Test & Method           & MCC\tabularnewline
\hline
Test 1 & HOG(10x10)    & 0.61 \tabularnewline
\hline
Test 2 & ULBP(8,1-8,8) & 0.60 \tabularnewline
\hline
Test 3 & Fusion         & 0.62 \tabularnewline
\hline
Test 4 & ULBP(8,5)     & 0.54 \tabularnewline
\hline
Test 5 & HOG(10x10)    & 0.59 \tabularnewline
\hline
Test 6 & ULBP(8,1-8,8) & \textbf{0.64} \tabularnewline
\hline
\end{tabular}
\end{table}

\subsection{Feature Relevance}\label{Features}

In this paper, we achieved a correct classification rate for the left and the right eyes of 82.83\% and 85.74\% with Non-occluded images and 80.00\% and 84.33\% with Occluded images. The best result obtained when using a combination of features was 83.60\%. When we merged all of the features extracted (raw pixels, shape, texture) with the same scale did not improve the results using texture or shape alone. Therefore, we estimated the histogram from pixels and concatenated with the histograms of ULBPs and HOGs to improve the results and reduces the length of feature vectors. Overall, the raw pixel produced the worst results.

As we have shown, the iris information is not relevant to classify gender when we used periocular images. Therefore, the most relevant feature for the classification should be located on the surrounded area of the iris (not in the iris). In order to demonstrate our statement, we estimate the relevance of each feature using the Gini Index with the XgBoost algorithm \cite{ChenGuestrin2016}. We tested several threshold values until we achieved the best results which were the correct classification rate of 89.22\% when using 4,900 features. This is shown in Table \ref{tab-xb}.

The relevance of the features was calculated explicitly for each attribute in the dataset. The attributes are ranked and compared to each other. In order to estimate the group of most relevant features, we use a decision tree. In the decision tree, we split the groups by using the relevance information weighted by the number of observations in the node. This differ from traditional decision tree methods where the threshold is used instead of relevance.

The performance was evaluated using the purity measure (Gini index). We select the split points to compute the relevance of the tree by averaging across it. See Table \ref{tab-xb}.
 
In Figure \ref{sietexg} we show an example of an image where the most relevant features located with the XgBoost algorithm are highlighted. This image was randomly selected from the periocular UND database. Each image has 19,200 pixels ($160\times120$ ) and shows the increment of the highlighted features from 500 to $10,000$. If we consider more features (pixels) than necessary, the redundancy of the information increase as shown in the figure \ref{sietexg} (d) with 10,000 features. 

Our initial experiments indicated that the relevant features for gender classification from periocular images are spread throughout the whole periocular area with exception of the iris. 
We validated our hypothesis that iris information contained in periocular images is less relevant than the information outside the iris for gender classification. The NIR images add extra information compared to VIS images. Thus, we can remove this area to improve the image-processing stage and gender classification. There are clear computational advantages to predicting gender using the same images like those in iris-recognition systems without needing other texture representations.

\begin{table*}[h]
\scriptsize
\centering
\caption{Features selected from Non occluded versus  occluded UND dataset using the best results with XgBoost algorithm from Table \ref{tab-und} and different thresholds values. N= Number of features selected, CCR= Correct Classification Rate.} \label{tab-xb}
\begin{tabular}{|c|c|c|c|c|c|c|c|c|c|c|c|}
\hline
\multicolumn{3}{|c|}{\textbf{LEFT EYE}}          & \multicolumn{3}{c|}{\textbf{RIGHT EYE}}         & \multicolumn{3}{c|}{\textbf{LEFT EYE}}          & \multicolumn{3}{c|}{\textbf{RIGHT EYE}}          \tabularnewline
\hline
\multicolumn{6}{|c|}{\textbf{Non Occluded}}                                                        & \multicolumn{6}{c|}{\textbf{Occluded}}                                                             \tabularnewline
\hline 
\textbf{Thresh.} & \textbf{N}   & \textbf{CCR.\%}  & \textbf{Thresh.} & \textbf{N}  & \textbf{CCR.\%}  & \textbf{Thresh.} & \textbf{N}  & \textbf{CCR.\%}  & \textbf{Thresh.} & \textbf{N}  & \textbf{CCR.\%}  \tabularnewline
\hline
0.00000          & 9,000          & 81.27          & 0.00000          & 9,000         & 85.94          & 0.00000          & 9,000         & 78.41          & 0.00000          & 9,000          & 84.35          \tabularnewline
\hline
0.00151          & 2,900          & 81.27          & 0.00152          & 2,900         & 85.94          & 0.00156          & 2,900         & 78.41          & \textbf{0.00148} & \textbf{3,020} & \textbf{84.35} \tabularnewline
\hline
0.00452          & 7,900           & 82.22          & 0.00759          & 2,900          & 85.94          & 0.01252          & 1,100          & 80.95          & 0.00890          & 2,100           & 79.55          \tabularnewline
\hline
0.00754          & 2,800           & 81.90          & 0.01366          & 1,000          & 80.83          & \textbf{0.00782} & \textbf{3,200} & \textbf{81.59} & 0.00742          & 3,200           & 81.15          \tabularnewline
\hline
\textbf{0.00302} & \textbf{1,470} & \textbf{83.90} & 0.00303          & 1,290         & 86.58          & 0.00313          & 1,400         & 79.37          & 0.01039          & 1,900           & 80.51          \tabularnewline
\hline
0.01056          & 1,500           & 78.41          & \textbf{0.0067}  & \textbf{4,900} & \textbf{89.22} & 0.00469          & 8,200          & 80.00          & 0.00593          & 5,000           & 82.75          \tabularnewline
\hline
0.01207          & 8,000            & 76.83          & 0.00455          & 8,100          & 87.22          & 0.01095          & 1,600          & 81.00          & 0.00455          & 7,200           & 83.07          \tabularnewline
\hline
0.00603          & 5,300           & 70.79          & 0.00910          & 1,900          & 84.35          & 0.00939          & 2,100          & 79.41          & 0.00297          & 1,300          & 83.39          \tabularnewline
\hline
\end{tabular}
\end{table*}

\subsection{Statistical Test}\label{ANOVA}

We employed an ANOVA test for functional data \cite{Cuevas2004} to analyze the results of Tables \ref{tab-und},  \ref{tab-uti},  \ref{tab-ce} and \ref{tab-all}, and to confirm that the differences in gender-classification rates of the predictive models, between Non occluded and Occluded periocular images, are not statistically significant. To get this, for each database we performed the experiment five times through a 5-fold cross-validation, and the average correct classification rate results were submitted to the ANOVA test. We applied this method to the results for different periocular images, from the left and right eyes, and also for different features extracted, from the Non occluded and Occluded iris images.

One of the advantages of the ANOVA procedure for functional data is that it allows us to deal with the multiplicity problem, which arises when a large number of statistical test are performed simultaneously. In this case, if decisions about the individual hypotheses are based on the unadjusted marginal p-values, then there is usually a high probability that some of the true null hypotheses will be rejected. 

For each experiment, the correct classification rates of different extracted features are modeled as functions, which are compared through the norm:

\begin{equation}
\|x\|:=\bigg( \int x^{2}(t)dt \bigg)^{1/2} =\bigg( \sum_{i=1}^{L}x_{i}^{2} \bigg)^{1/2}, 
\end{equation}
where $L$ is the number of features considered and $x_{i}$ is the associated correct classification rate.

The p-values of Cuevas et al. \cite{Cuevas2004} procedure are calculated in terms of the bootstrap simulations where the null hypothesis is $H_{0}$: the mean precision functions between Non occluded and Occluded images are equal''. A p-value $p\leq 0.05$ indicates strong evidence against the null hypothesis, so we can reject the null hypothesis. A p-value $p>0.05$ shows weak evidence against the null hypothesis, so we fail to reject the null hypothesis. Since the p-values $0.1789$, $0.2146$ and $0.1675$ are not lower than the significance level $\alpha= 5\%$, then there is no statistical evidence that there exists a difference in the mean correct classification rate obtained in the gender-classification using Non occluded and Occluded iris images. See Figure \ref{figure:1}, \ref{figure:2} and \ref{figure:3} in appendix \ref{sec14}.

\section{Conclusions}

In this paper we presented competitive gender-classification results with respect to the state of the art. Even when using several databases captured in different conditions. Our approach does not need the image to be centered by the pupil.

We showed that periocular features have high quality information for classifying gender. The results obtained for classification are similar to other approaches even-though we use fewer features. Most traditional methods use several stages such as acquisition, segmentation, normalization and encoding information while we only use the acquisition stage.  We also show the high redundancy (not useful information) present in the images. Therefore, it is possible to reduce the number of features (pixels) using the XgBoost algorithm and to select the most relevance features.

The highest correct classification rate reported in previous works was 85.7\% \cite{BobeldykRoss2016}. Also, Kuehlkamp et al.  approach \cite{Kuehlkamp-2019}, complement our result and to validate our conclusion with the analysis of the automatic feature extraction method using deep learning methods. These results confirm that periocular information reaches better results.

In this paper, we achieve a correct classification rate of  89.22\% when  using the Xgboost algorithm with only 4,900 features and a threshold of 0.00607 for the right eye. However, our results were tested on databases that were not used for training the best parameters. This shows the robustness of our method. Our approach is fully automatic and does not require initialization as state of the art techniques. We use a circular segmentation mask instead of the square one used in \cite{BobeldykRoss2016}. This allows us to better represent the iris shape. 

In order to evaluate the results, we included two new databases: Periocular UND database (a subset manually labeled for gender-classification and automatically segmented and masked) and UNAB-Gender a homemade database. Furthermore, we presented results that showed comparable correct classification rate for NIR disjointed with gender-from-periocular databases using the same images as those used in iris-recognition systems. This contribution suggests that Non occluded and Occluded information are very similar, and therefore the NIR information obtained from the iris does not improve gender classification. Gender can be classified with periocular NIR images without the need to include iris information. 

The cross-performance analysis using different datasets with different image conditions and captured with different sensors, shows that shape and texture information is more relevant than intensity information. Therefore, we can reduce the amount of information required by the classifiers (SVM and ensemble methods). The concatenation of the texture feature at different scales (instead of using only one) from the same image such as ULBP improves the gender classification. 

The XgBoost algorithm was used to select the most relevant features of the periocular image. This allows us to significantly reduce the number of features selected by only using the most relevant ones. This improves the gender-classification rates.

\section{Future work}
The iris information is an unique pattern have been shown to be stable over the years. This still can be not established for periocular images because of using some skin information areas. The skin changes with years and some wrinkles appear. Therefore, much more work must be done before establishing a general conclusion even create a new dataset to be able to measure these changes and the validation of the results. This work going in that direction.

\section{Acknowledgements}
This work was partially supported by the Universidad Tecnologica de Chile - INACAP, Universidad Andres Bello - DCI, and FONDECYT INICIACION N11170189.

\bibliographystyle{unsrt}
\bibliography{References_peri}

\section{Appendix}\label{sec14}
\vspace{-0.5cm}
\begin{figure*}[h]
\begin{centering}
\begin{tabular}{c}
{\includegraphics[scale=0.24]{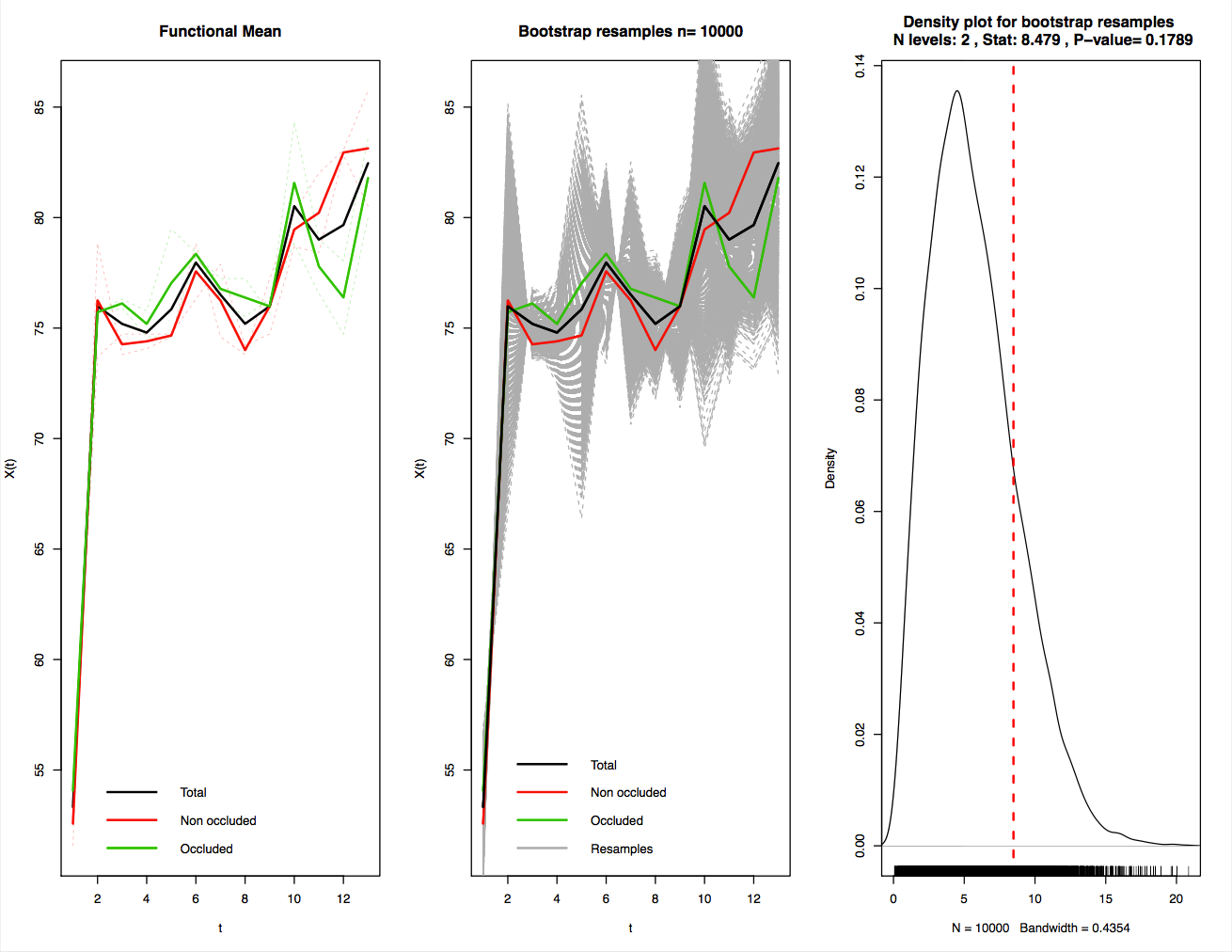}}
\end{tabular}
\par\end{centering}
\caption{ANOVA test for functional data for periocular images trained and tested using \textbf{UND Periocular}. Left side: correct classification rate functions for the different features studied. From 1 up to 13: Raw, ULBP 8,1; ULBP 8,2; ULBP 8,3; ULBP 8,4; ULBP 8,5; ULBP 8,6; ULBP 8,7; ULBP 8,8; ULBP 8,1 to 8,8; HOG 3 x 3; HOG 5 x 5; HOG 10 x 10 for occluded images (Green), Non-occluded (Red) and their total average (Black). At the center, bootstrap simulations of the average precision functions are shown. On the right is the kernel estimator of the density of the statistic proposed by (see Cuevas et al., 2004) for the computation of the p-value.}
\label{figure:1}
\end{figure*}

\begin{figure*}[h]
\begin{centering}
\begin{tabular}{c}
{\includegraphics[scale=0.24]{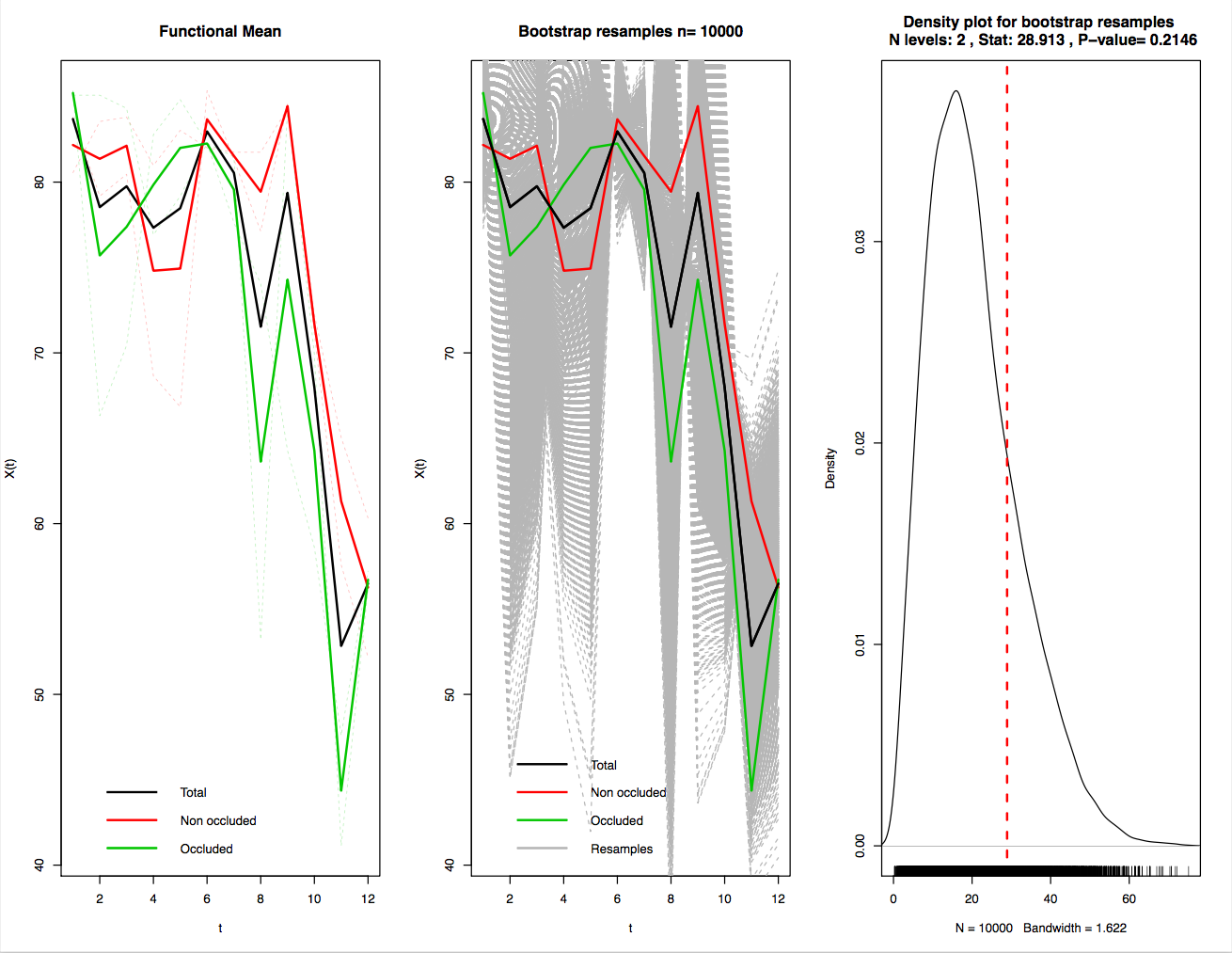}}
\end{tabular}
\par\end{centering}
\caption{ANOVA test for functional data for periocular images trained with \textbf{UND Periocular} and tested with \textbf{UTIRIS NIR}. Left side: the correct classification rate functions for the different features studied. From 1 up to 12: ULBP 8,1; ULBP 8,2; ULBP 8,3; ULBP 8,4; ULBP 8,5; ULBP 8,6; ULBP 8,7; ULBP 8,8; ULBP 8,1 to 8,8; HOG 3 x 3; HOG 5 x 5; HOG 10 x 10 for occluded images (Green), Non-occluded (Red) and their total average (Black). At the center, bootstrap simulations of the average precision functions are shown. On the right is the kernel estimator of the density of the statistic proposed by [see Cuevas et al., 2004] for the computation of the p-value.}
\label{figure:2}
\end{figure*}

\begin{figure*}[h]
\begin{centering}
\begin{tabular}{c}
{\includegraphics[scale=0.25]{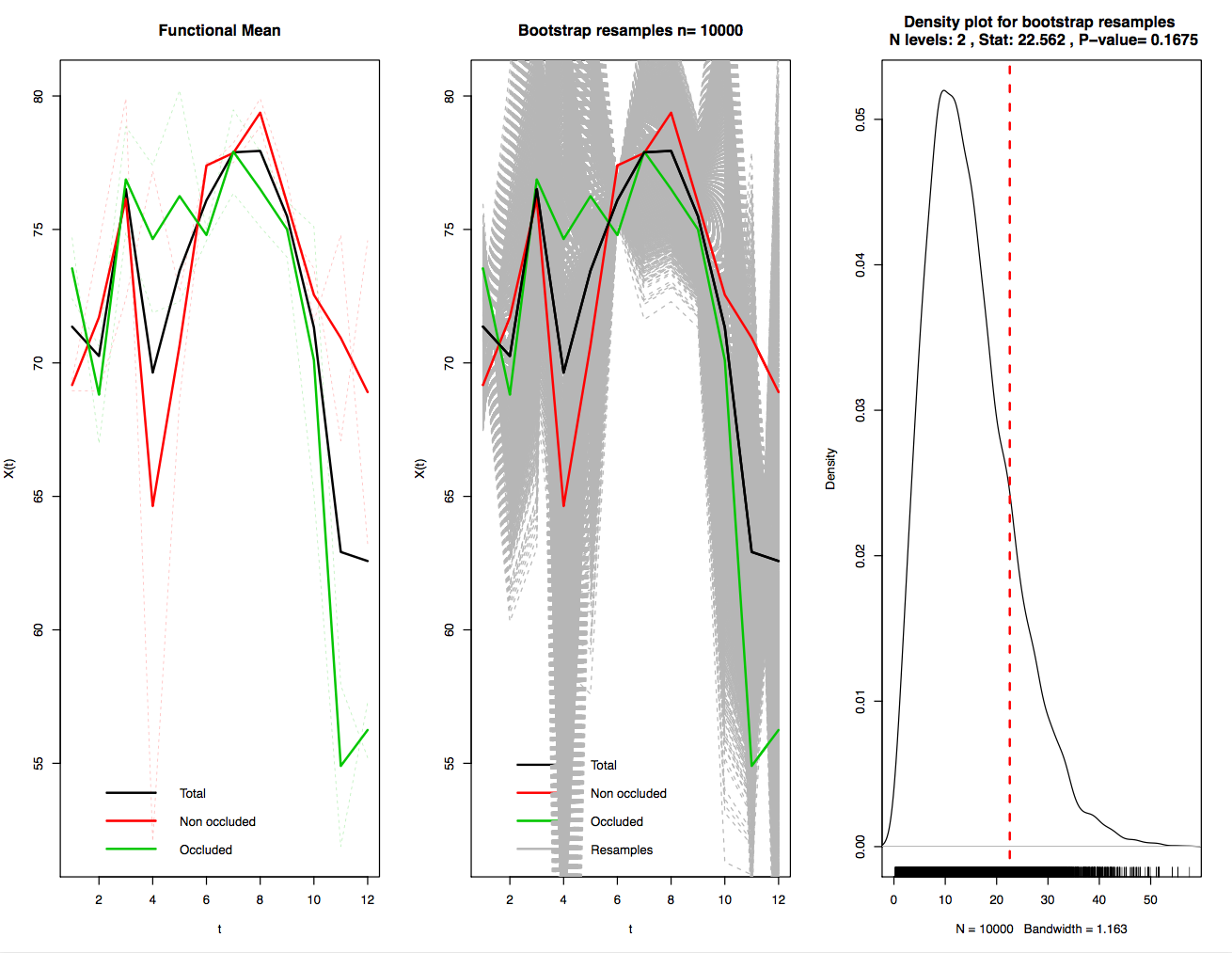}}
\end{tabular}
\par\end{centering}
\caption{ANOVA test for functional data for periocular images trained with \textbf{UND Periocular} and tested with \textbf{Cross-Eye-2016 NIR}. Left side: the correct classification rate functions for the different features studied. From 1 up to 12: ULBP 8,1; ULBP 8,2; ULBP 8,3; ULBP 8,4; ULBP 8,5; ULBP 8,6; ULBP 8,7; ULBP 8,8; ULBP 8,1 to 8,8; HOG 3 x 3; HOG 5 x 5; HOG 10 x 10 for occluded images (Green), Non-occluded (Red) and their total average (Black). At the center, bootstrap simulations of the average precision functions are shown. On the right is the kernel estimator of the density of the statistic proposed by [see Cuevas et al., 2004] for the computation of the p-value.}
\label{figure:3}
\end{figure*}

\end{document}